\documentclass{article}
\usepackage{ijcai19}

\usepackage{times}
\usepackage{soul}
\usepackage{url}
\usepackage[hidelinks]{hyperref}
\usepackage[utf8]{inputenc}
\usepackage[small]{caption}
\usepackage{graphicx}
\usepackage{amsmath}
\usepackage{booktabs}
\usepackage{algorithm}
\usepackage{algorithmic}
\usepackage{times}
\usepackage{latexsym}
\usepackage{multirow}
\usepackage{CJK} 
\usepackage{pinyin}
\usepackage{color}
\usepackage{booktabs}
\usepackage[normalem]{ulem}
\usepackage{amssymb}
\usepackage{arydshln}

\title{Sequence Generation: From Both Sides to the Middle}

\author{
Long Zhou$^{1,2}$
\and
Jiajun Zhang$^{1,2}$\and
Chengqing Zong$^{1,2,3}$\And
Heng Yu$^4$
\affiliations
$^1$University of Chinese Academy of Sciences, Beijing, China\\
$^2$National Laboratory of Pattern Recognition, CASIA, Beijing, China\\
$^3$CAS Center for Excellence in Brain Science and Intelligence Technology, Shanghai, China\\
$^4$Machine Intelligence Technology Lab, Alibaba Group
\emails
\{long.zhou, jjzhang, cqzong\}@nlpr.ia.ac.cn,
yuheng.yh@alibaba-inc.com
}

\begin{document}
\begin{CJK*}{UTF8}{gkai}

\maketitle

\begin{abstract}
The encoder-decoder framework has achieved promising process for many sequence generation tasks, such as neural machine translation and text summarization. Such a framework usually generates a sequence token by token from left to right, hence (1) this autoregressive decoding procedure is time-consuming when the output sentence becomes longer, and (2) it lacks the guidance of future context which is crucial to avoid under-translation.
To alleviate these issues, we propose a synchronous bidirectional sequence generation (SBSG) model which predicts its outputs from both sides to the middle simultaneously.
In the SBSG model, we enable the left-to-right (L2R) and right-to-left (R2L) generation to help and interact with each other by leveraging interactive bidirectional attention network.
Experiments on neural machine translation (En$\Rightarrow$De, Ch$\Rightarrow$En, and En$\Rightarrow$Ro) and text summarization tasks show that the proposed model significantly speeds up decoding while improving the generation quality compared to the autoregressive Transformer.
\end{abstract}


\section{Introduction}

The neural encoder-decoder framework has been widely adopted in sequence generation tasks, including neural machine translation (NMT)~\cite{Sutskever:2014,Bahdanau:2015,vaswani2017attention}, text summarization~\cite{D15-1044,P17-1101,li-etal-2018-ensure}, and image captioning~\cite{Xu2015Show,Vinyals2015Show}.
In this framework, the encoder models the semantics of the input sentence and transforms it into a context vector representation, which is then fed into the decoder to generate the output sequence token by token in a left-to-right manner.

\begin{figure}
	\setlength{\abovecaptionskip}{+0.15cm}
	\setlength{\belowcaptionskip}{-0.2cm}
	\centering
	\includegraphics[width=7.5cm]{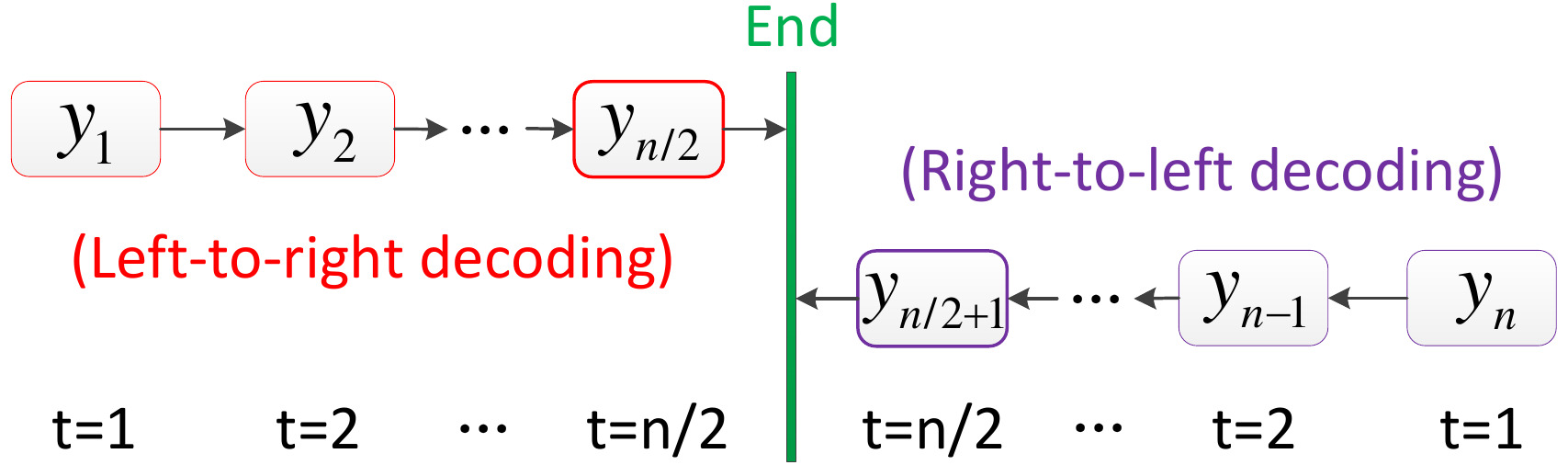}
	\caption{The illustration of synchronous bidirectional decoding for sequence-to-sequence model.
		The bidirectional decoder, predicting its outputs from left to right and from right to left simultaneously and interactively, can produce two tokens at each time step.
	}\label{Bi-ge}
\end{figure}

Although the framework has obtained great success, the sequence-to-sequence model suffers from the decoding efficiency problem~\cite{gu2017non}.
Most of the models use autoregressive decoders that operate one step at a time, and they become slow when generating long sequences because a computationally intensive neural network is used to predict each token.
Several recently proposed models avoid recurrence at training time by leveraging convolutions~\cite{gehring2017convolutional} or self-attention~\cite{vaswani2017attention} as more-parallelizable alternatives to recurrent neural networks, but the decoding process cannot share the speed strength of parallelization due to the autoregressive generation schema in the decoder.
More importantly, this left-to-right decoding cannot take advantage of future contexts which can be generated in a right-to-left decoding~\cite{zhang2018asynchronous}.

To avoid this autoregressive property, 
\citeauthor{gu2017non}~\shortcite{gu2017non} proposed a non-autoregressive model to speed up machine translation by directly generating target words without relying on any previous predictions.
\citeauthor{oord2017parallel}~\shortcite{oord2017parallel} modified a convolutional network for non-autoregressive modeling of speech synthesis.
\citeauthor{lee2018deterministic}~\shortcite{lee2018deterministic} introduced a conditional non-autoregressive neural sequence model based on iterative refinement.
However, in spite of their improvement in decoding speed, non-autoregressive models typically 
suffer from the substantial drop in generation quality.

In this paper, we propose a synchronous bidirectional sequence generation (SBSG) model to achieve a better improvement on both generation quality and decoding speed.
Instead of producing output sentences token by token or predicting its outputs in a totally parallel manner, the SBSG model generates two tokens at a time.
As shown in Figure~\ref{Bi-ge}, the bidirectional decoder can generate output sentences from both sides to the middle with both left-to-right (L2R) and right-to-left (R2L) directions.
Furthermore, we introduce an interactive bidirectional attention network to bridge L2R and R2L outputs.
More specifically, at each moment, the generation of target-side tokens does not only rely on its previously generated outputs (history information), but also depends on previously predicted tokens of the other generation direction (future information).

Specifically, the contributions of this paper can be summarized as two folds:
\begin{itemize}
	\item We propose a novel SBSG model that employs one decoder to predict outputs from both sides to the middle simultaneously and interactively. 
	To the best of our knowledge, this is the first work to perform sequence generation from both ends to the middle.
	\item We extensively evaluate the proposed model on typical sequence generation tasks, namely neural machine translation and text summarization. 
	In the case of \textbf{machine translation}, we not only obtain approximately 1.4$\times$ (1.5$\times$) speedup for decoding than autoregressive Transformer with beam search (greedy search), but also get an improvement of 0.39 (0.99), 1.26 (2.87) and 0.73 (1.11) BLEU points of translation quality in WMT14 En$\Rightarrow$De, NIST Ch$\Rightarrow$En and WMT16 En$\Rightarrow$Ro respectively, which also significantly outperforms previous non-autoregressive models~\cite{gu2017non,lee2018deterministic,kaiser2018fast}.
	For \textbf{text summarization}, the proposed model is able to decode approximately 1.5$\times$ faster while achieving better generation quality relative to the autoregressive counterparts. 
	
\end{itemize}

\section{Related Work}

\paragraph{Autoregressive Decoding.}
Recent approaches to sequence to sequence learning typically leverage recurrence ~\cite{Sutskever:2014}, convolution ~\cite{gehring2017convolutional}, or self-attention ~\cite{vaswani2017attention} as basic building blocks.
Particularly, relying entirely on the attention mechanism, the Transformer introduced by \citeauthor{vaswani2017attention}~\shortcite{vaswani2017attention} can improve the training speed as well as model performance.
To accelerate autoregressive architecture, \citeauthor{P16-2021}~\shortcite{P16-2021} introduced a sentence-level vocabulary which is able to reduce computing time and memory usage.
\citeauthor{D17-1300}~\shortcite{D17-1300} focused on fast and accurate neural machine translation decoding in CPU. 
\citeauthor{P18-1166}~\shortcite{P18-1166} proposed an average attention network (AAN) as an alternative to the self-attention network in the decoder of Transformer. 
Despite their remarkable success, they are difficult to parallelize and this unidirectional decoding framework limits its potential~\cite{liu2016agreementb}.

\paragraph{Non-Autoregressive Decoding.}
In terms of speeding up the decoding of the neural Transformer,
\citeauthor{gu2017non}~\shortcite{gu2017non} modified the autoregressive architecture to speed up machine translation by directly generating target words in parallel.
However, the main drawback of this work is the need for extensive policy gradient fine-turning techniques, as well as the issue that this method only works for machine translation and cannot be applied to other sequence generation tasks.
In parallel to \citeauthor{gu2017non}~\shortcite{gu2017non}, \citeauthor{oord2017parallel}~\shortcite{oord2017parallel} presented a successful, non-autoregressive sequence model for speech waveform. 
Besides, \citeauthor{kaiser2018fast}~\shortcite{kaiser2018fast} first auto-encoded the target sequence into a shorter sequence of discrete latent variables, and then decoded the output sentence from this shorter latent sequence in parallel.
\citeauthor{lee2018deterministic}~\shortcite{lee2018deterministic} introduced a conditional non-autoregressive neural sequence model based on iterative refinement.
Concurrently to our work, \citeauthor{wang2018semi}~\shortcite{wang2018semi} presented a semi-autoregressive Transformer for faster translation without changing the autoregressive property in global.
However, these approaches improved the parallelizability but significantly reduced  generation quality.

\paragraph{Towards Bidirectional Decoding.}
\citeauthor{liu2016agreementb}~\shortcite{liu2016agreementb} proposed an agreement model to encourage the agreement between a pair of target-directional LSTMs, which generated more balanced targets. 
Similarly, some work attempted at target-bidirectional decoding for SMT or NMT~\cite{watanabe2002bidirectional,D09-1117,liu2016agreementa,W16-2323,liu2018}.
Recently, \citeauthor{zhang2018asynchronous}~\shortcite{zhang2018asynchronous} and \citeauthor{zhou2019synchronous}~\shortcite{zhou2019synchronous} proposed an asynchronous and synchronous bidirectional decoding for NMT, respectively.
\citeauthor{serdyuk2018twin}~\shortcite{serdyuk2018twin} presented the twin networks to encourage the hidden state of the forward network to be close to that of the backward network used to predict the same token.
Nevertheless, the above studies are not to speed up the decoding procedure, and even sacrifice speed in exchange for quality improvement.
Our work differs from those by introducing a novel sequence generation model which aims at taking full advantage of both left-to-right and right-to-left decoding to accelerate and improve sequence generation.

\section{The Framework}

\begin{figure}
	\setlength{\abovecaptionskip}{+0.15cm}
	\setlength{\belowcaptionskip}{-0.2cm}
	\centering
	\includegraphics[width=8.5cm]{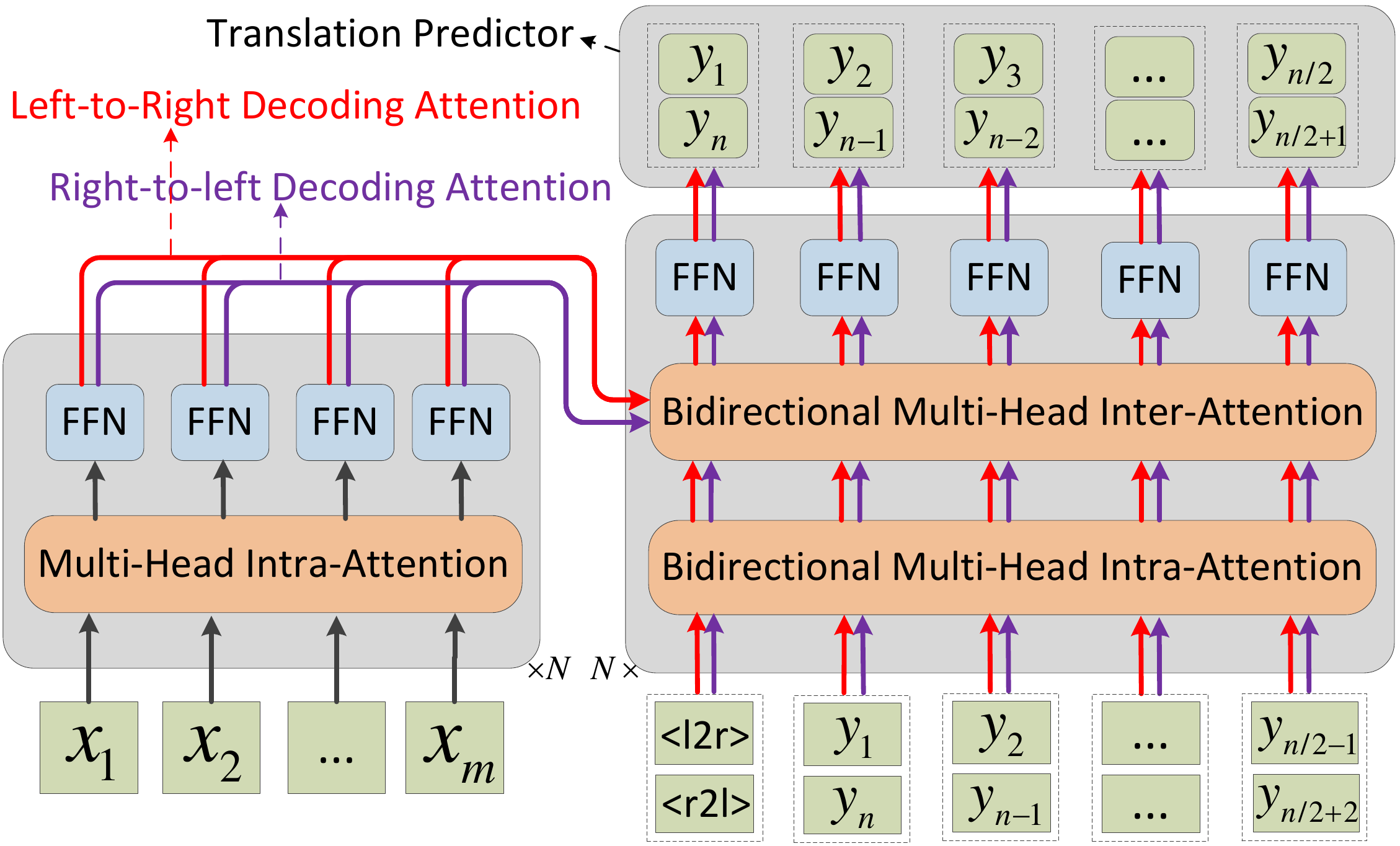}
	\caption{The new Transformer architecture with the proposed bidirectional multi-head (intra- and inter-) attention network. 
		Instead of producing output sentence token by token or predicting its outputs in totally parallel, the proposed model generates two tokens (one from left to right, the other one from right to left) at a time, indicated by two special labels.
	}\label{Bi-trans}
\end{figure}

Our goal in this work is to achieve a better improvement on both generation quality and decoding speed.
We introduce a novel method for decoding with both left-to-right and right-to-left manners simultaneously and interactively in a unified model.
As demonstrated in Figure~\ref{Bi-trans}, our proposed model consists of an encoder and a bidirectional decoder, in which two special labels ($\langle l2r \rangle$ and $\langle r2l \rangle$) at the beginning of output sentence are utilized to guide the sequence generation from left to right or right to left.
The bidirectional decoder reads the encoder representation and generates two output tokens at each time step, by using interactive bidirectional attention networks.
Next, we will detail individual components and introduce an algorithm for training and inference.

\subsection{The Neural Encoder}

Given an input sentence $x = {(x_1, x_2, ..., x_m)}$, the new Transformer leverages its encoder to induce input-side semantic and dependencies so as to enable its decoder to recover the encoded information in an output sentence. 
The encoder is composed of a stack of N identical layers, each of which has two sub-layers:
\begin{equation}
\small
\setlength{\abovedisplayskip}{3pt}
\setlength{\belowdisplayskip}{3pt}
	\begin{aligned}
		\widetilde{h}^l &= {\rm{LN}}(h^{l-1} + {\rm{MHAtt}}(h^{l-1}, h^{l-1}, h^{l-1})) \\
		h^l &= {\rm{LN}}(\widetilde{h}^l + {\rm{FFN}}(\widetilde{h}^l))
	\end{aligned}
\end{equation}
where the superscript $l$ indicates layer depth, LN is layer normalization, FFN means feed-forward networks, and MHAtt denotes the multi-head attention mechanism as follows.

\paragraph{Scaled Dot-Product Attention.}
An attention function can be described as mapping a query and a set of key-value pairs to an output. The output is computed as a weighted sum of the values, where the weight assigned to each value is computed by a compatibility function of the query and the corresponding key. 
Scaled dot-product attention operates on a query $Q$, key $K$, and a value $V$ as:
\begin{equation}
\small
\setlength{\abovedisplayskip}{3pt}
\setlength{\belowdisplayskip}{3pt}
	{\rm{ATT}}(Q,K,V) = {\rm{softmax}}(\frac{QK^T}{\sqrt{d_k}})  V   \label{attention}
\end{equation}
where $d_k$ is the dimension of the key.

\paragraph{Multi-Head Attention.}
We use the multi-head version with $h$ heads. It obtains $h$ different representations of ($Q,K,V$), computes scaled dot-product attention for each representation, concatenates the results, and projects the concatenation with a feed-forward layer. 
\begin{equation}
\small
\setlength{\abovedisplayskip}{3pt}
\setlength{\belowdisplayskip}{3pt}
	\begin{aligned} \label{MHAtt}
		{\rm{MHAtt}}(Q,K,V) = {\rm{Concat}}({\rm{head}}_1,...,{\rm{head}}_h)W^O  \\ 
		{\rm{head}}_i = {\rm{ATT}}(QW_i^Q, KW_i^K, VW_i^V)  
	\end{aligned}
\end{equation}
where  $W_i^Q$, $W_i^K$, $W_i^V$ and $W^O$ are parameter matrices.

\subsection{The Bidirectional Decoder}

The bidirectional decoder performs decoding in both left-to-right and right-to-left manners under the guidance of previously generated forward and backward outputs. We apply our bidirectional attention network to replace the self-attention network in its decoder part, and illustrate the overall architecture in Figure~\ref{Bi-trans}.
Next, we will present those two bidirectional attention models and integrate them into the decoder of Transformer.

\subsubsection{Bidirectional Scaled Dot-Product Attention}

\begin{figure}
	\setlength{\abovecaptionskip}{+0.15cm}
	\setlength{\belowcaptionskip}{-0.2cm}
	\centering
	\includegraphics[width=8cm]{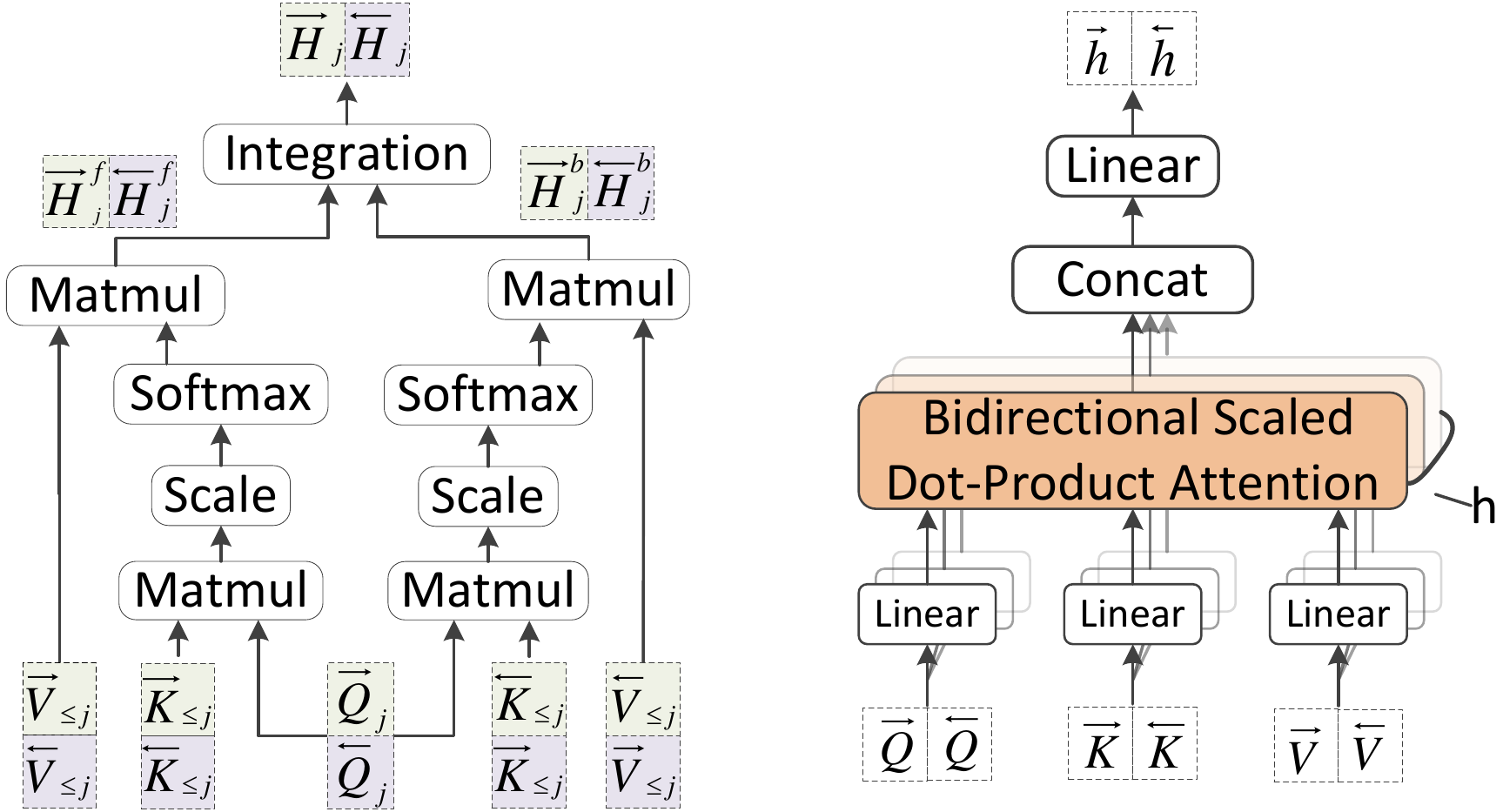}
	\caption{(left) Bidirectional Scaled Dot-Product Attention operates on forward (L2R) and backward (R2L) queries $Q$, keys $K$, values $V$. (right) Bidirectional Multi-Head Intra-Attention consists of several attention layers in parallel.
	}\label{Bi-att}
\end{figure}

Figure~\ref{Bi-att} (left) shows our particular attention.
The input consists of queries ([$\overrightarrow{Q}$;$\overleftarrow{Q}$]), keys ([$\overrightarrow{K}$;$\overleftarrow{K}$]) and values ([$\overrightarrow{V}$;$\overleftarrow{V}$])
which are all concatenated by forward (L2R) states and backward (R2L) states.
The new forward states $\overrightarrow{H}_{j}$ and backward states $\overleftarrow{H}_{j}$ can be obtained by bidirectional dot-product scaled attention. 
For new forward states $\overrightarrow{H}_{j}$, it can be calculated as:
\begin{equation}\label{F-SBA}
\small
\setlength{\abovedisplayskip}{0cm}
\setlength{\belowdisplayskip}{0.1cm}
	\begin{aligned}
		\overrightarrow{H}_{j}^{f} &= {\rm{ATT}}(\overrightarrow{Q}_{j}, \overrightarrow{K}_{\leq j}, \overrightarrow{V}_{\leq j}) 
		= {\rm{softmax}}(\frac{\overrightarrow{Q}_{j} \overrightarrow{K}_{\leq j}^T}{\sqrt{d_k}})  \overrightarrow{V}_{\leq j} \\
		\overrightarrow{H}_{j}^{b} &= {\rm{ATT}}(\overrightarrow{Q}_{j}, \overleftarrow{K}_{\leq j}, \overleftarrow{V}_{\leq j}) 
		= {\rm{softmax}}(\frac{\overrightarrow{Q}_{j} \overleftarrow{K}_{\leq j}^T}{\sqrt{d_k}})  \overleftarrow{V}_{\leq j} \\
	\end{aligned}
\end{equation}
where $\overrightarrow{H}_{j}^{f}$ is obtained by conventional scaled dot-product attention as introduced in Equation~\ref{attention},
and $\overrightarrow{H}_{j}^{b}$ contains the attentional future information from R2L decoding.
Then we use a linear interpolation method to integrate the forward information $\overrightarrow{H}_{j}^{f}$ and backward information $\overrightarrow{H}_{j}^{b}$:
\begin{equation}\label{F-SBA-2}
\small
	\setlength{\abovedisplayskip}{3pt}
	\setlength{\belowdisplayskip}{3pt}
	\begin{aligned}
	\overrightarrow{H}_{j} &= {\rm{Integration}}(\overrightarrow{H}_{j}^{f}, \overrightarrow{H}_{j}^{b}) = \overrightarrow{H}_{j}^{f} + \lambda * \overrightarrow{H}_{j}^{b}
	\end{aligned}
\end{equation}
where $\lambda$ is a hyper-parameter decided by the performance on development set.

For R2L decoding, similar to the calculation of forward hidden states $\overrightarrow{H}_{j}$, the backward hidden states $\overleftarrow{H}_{j}$ can be computed as follows.
\begin{equation}\label{B-SBA}
\small
\setlength{\abovedisplayskip}{3pt}
\setlength{\belowdisplayskip}{3pt}
	\begin{aligned}
		\overleftarrow{H}_{j}^{f} &= {\rm{ATT}}(\overleftarrow{Q}_{j}, \overleftarrow{K}_{\leq j}, \overleftarrow{V}_{\leq j}) \\
		\overleftarrow{H}_{j}^{b} &= {\rm{ATT}}(\overleftarrow{Q}_{j}, \overrightarrow{K}_{\leq j}, \overrightarrow{V}_{\leq j}) \\
		\overleftarrow{H}_{j} &= {\rm{Integration}}(\overleftarrow{H}_{j}^{f}, \overleftarrow{H}_{j}^{b})
	\end{aligned}
\end{equation}
where Integration($\cdot$) is the same as introduced in Equation~\ref{F-SBA-2}. 
We refer to the whole procedure formulated in Equation~\ref{F-SBA}-\ref{B-SBA} as BSDPA($\cdot$).
\begin{equation}\label{BiAtt}
\small
\setlength{\abovedisplayskip}{3pt}
\setlength{\belowdisplayskip}{3pt}
	\begin{aligned}
		[][ \overrightarrow{H}_{j};\overleftarrow{H}_{j}] = \rm{BSDPA}([\overrightarrow{Q}_{j};\overleftarrow{Q}_{j}],[\overrightarrow{K}_{\leq j};\overleftarrow{K}_{\leq j}], [\overrightarrow{V}_{\leq j};\overleftarrow{V}_{\leq j}])
	\end{aligned}
\end{equation}
It is worth noting that $\overrightarrow{H}_{j}$ and $\overleftarrow{H}_{j}$ can improve each other and be calculated in parallel.

\subsubsection{Bidirectional Multi-Head Intra-Attention}

Different from the mask multi-head attention (Equation~\ref{MHAtt}), we can obtain the new forward and backward hidden states simultaneously, as shown in Figure~\ref{Bi-att} (right),
where $i$-th attention head with $j$-th target token can be computed using BSDPA($\cdot$):
\begin{equation}
\small
\setlength{\abovedisplayskip}{0.1cm}
\setlength{\belowdisplayskip}{0.1cm}
	\begin{aligned}
		{\rm{head}}_{i,j} = [\overrightarrow{h}_{i,j};\overleftarrow{h}_{i,j}] = {\rm{BSDPA}}([\overrightarrow{Q}_{j};\overleftarrow{Q}_{j}]{W_i^Q}, \\
		[\overrightarrow{K}_{\leq j};\overleftarrow{K}_{\leq j}]W_i^K, [\overrightarrow{V}_{\leq j};\overleftarrow{V}_{\leq j}]W_i^V)
	\end{aligned}
\end{equation}
where $W_i^Q$, $W_i^K$ and $W_i^V$ are parameter matrices, which are the same as standard multi-head attention introduced in Equation~\ref{MHAtt}.
By contrast, bidirectional multi-head inter-attention is composed of two standard multi-head attention models, which do not interact with each other.

\subsubsection{Integrating Bidirectional Attention into Decoder}

We use our bidirectional attention network to replace the multi-head attention in the decoder part, as demonstrated in Figure~\ref{Bi-trans}.
For each layer in bidirectional decoder, the first sub-layer is the \textbf{bidirectional multi-head intra-attention} ($\rm{BiAtt^{Intra}}$) network{\footnote[1]{Note that we follow \citeauthor{vaswani2017attention}~\cite{vaswani2017attention} to use residual connection and layer normalization in each decoder sub-layer, which are omitted in the presentation for simplicity.}} which is capable of combining history and future information:
\begin{equation}
\small
\setlength{\abovedisplayskip}{3pt}
\setlength{\belowdisplayskip}{3pt}
	\begin{aligned}
		[][\overrightarrow{s}^l_d;\overleftarrow{s}^l_d] =  {\rm{BiAtt^{Intra}}}([\overrightarrow{s}^{\tiny{l-1}};\overleftarrow{s}^{l-1}], [\overrightarrow{s}^{l-1};\overleftarrow{s}^{l-1}], \\ [\overrightarrow{s}^{l-1};\overleftarrow{s}^{l-1}])
	\end{aligned}
\end{equation}
where $s^l$ denotes $l$-layer hidden states or embedding vectors when $l$=0, and subscript $d$ denotes the decoder-informed intra-attention representation.

The second sub-layer is the \textbf{bidirectional multi-head inter-attention} ($\rm{BiAtt^{Inter}}$) which integrates the representation of the corresponding source sentence by performing left-to-right and right-to-left decoding attention respectively, as shown in Figure~\ref{Bi-trans}.
\begin{equation}
\small
\setlength{\abovedisplayskip}{3pt}
\setlength{\belowdisplayskip}{3pt}
	\begin{aligned}
		[][\overrightarrow{s}^l_e;\overleftarrow{s}^l_e] &= {\rm{BiAtt^{Inter}}}([\overrightarrow{s}^{l}_d;\overleftarrow{s}^{l}_d], [h^N;h^N], [h^N;h^N]) \\
	\end{aligned}
\end{equation}
where $e$ denotes the encoder-informed inter-attention representation, and $h^N$ is the source hidden state of top layer.

The third sub-layer is a position-wise fully connected feed-forward neural network: $[\overrightarrow{s}^l; \overleftarrow{s}^l] =  {\rm{FFN}}([\overrightarrow{s}^l_e;\overleftarrow{s}^l_e])$.

Finally, we employ a linear transformation and softmax activation to compute the probability of the $j$-th tokens based on $N$-layer $s_j^N=[\overrightarrow{s}_j^N;\overleftarrow{s}_j^N]$, namely the final hidden states of forward and backward decoding.
\begin{equation}
\small
\setlength{\abovedisplayskip}{3pt}
\setlength{\belowdisplayskip}{3pt}
	\begin{aligned}
		p(\overrightarrow{y}_{j}|\overrightarrow{y}_{<j},\overleftarrow{y}_{<j},x,\theta) = {\rm{softmax}}(\overrightarrow{s}_j^NW)  \\
		p(\overleftarrow{y}_{j}|\overleftarrow{y}_{<j},\overrightarrow{y}_{<j},x,\theta) = {\rm{softmax}}(\overleftarrow{s}_j^NW)
	\end{aligned}
\end{equation}
where $W$ denotes the weight matrix and $\theta$ is the shared parameters for L2R and R2L decoding.

\subsection{Training and Inference}

\begin{figure}[b]
	\vspace{-0.3cm}
	\setlength{\abovecaptionskip}{+0.15cm}
	\centering
	\includegraphics[width=7.0cm]{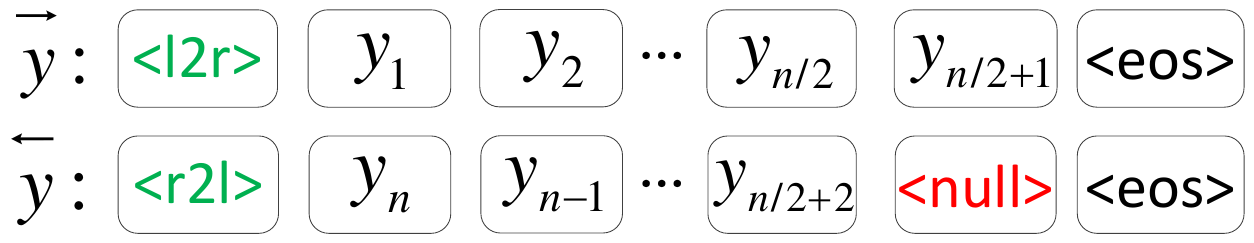}
	\caption{The smoothing model introduced to connect L2R and R2L results smoothly. When the output sentence has odd tokens, we randomly insert $\langle null \rangle$ which means null word and can be removed in postprocessing.
	}\label{smooth}
\end{figure}

\paragraph{Training.}
Given a parallel sentence pair ($x$, $y$), we design a smart strategy to enable synchronous bidirectional generation within a decoder.
We first divide the output sentence ($y$) into two halves
and reverse the second half.
Second, we separately add the special labels ($\langle l2r \rangle$ and $\langle r2l \rangle$) at the beginning of each half sentence ($\overrightarrow{y}$ and $\overleftarrow{y}$) to guide generating tokens from left to right or right to left.
Finally, we propose a smoothing model to better connect both directional generational results. As shown in Figure~\ref{smooth}, if the output length is odd, we add the additional tag ($\langle null \rangle$) before $\langle eos \rangle$ in forward or backward sentence randomly.
In other words, our model is capable of generating a null word when necessary.
Following previous work~\cite{gu2017non,wang2018semi}, we also use knowledge distillation techniques~\cite{kim2016sequence} to train our model.
Given a set of training examples $\{x^{(z)}, y^{(z)}\}^Z_{z=1}$, the training algorithm aims to find the model parameters that maximize the likelihood of the training data:
\begin{equation}
\small
\setlength{\abovedisplayskip}{+0.05cm}
\setlength{\belowdisplayskip}{-0.1cm}
	\begin{aligned}
		\mathbb{L}(\theta) = \frac{1}{Z} \sum_{z=1}^Z \sum_{j=1}^{n/2} \{ log \ p(\overrightarrow{y}^{(z)}_{j}|\overrightarrow{y}^{(z)}_{<j},\overleftarrow{y}^{(z)}_{<j},x^{(z)},\theta) \\
		+ \ log \ p(\overleftarrow{y}^{(z)}_{j}|\overleftarrow{y}^{(z)}_{<j}, \overrightarrow{y}^{(z)}_{<j}, x^{(z)},\theta)\}
	\end{aligned}
\end{equation}
\begin{figure}
	\setlength{\abovecaptionskip}{+0.15cm}
	\setlength{\belowcaptionskip}{-0.2cm}
	\centering
	\includegraphics[width=7.5cm]{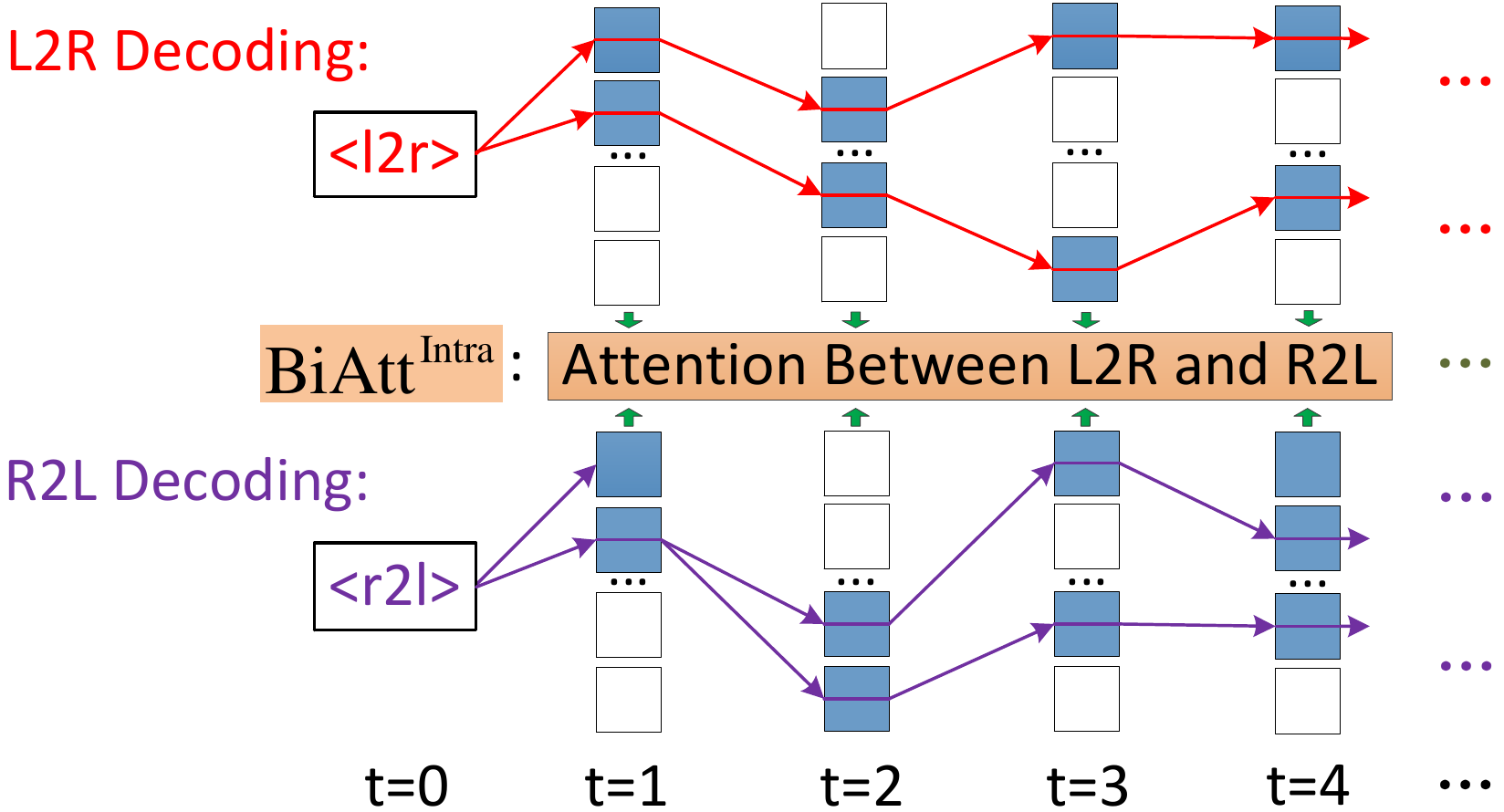}
	\caption{The bidirectional beam search process of our proposed model, which produces tokens from left-to-right and right-to-left simultaneously, under the guidance of two special labels ($\langle${l2r}$\rangle$ and $\langle${r2l}$\rangle$). By using bidirectional attention model, left-to-right and right-to-left decoding can help and interact with each other.
	}\label{decoding}
\end{figure}

\paragraph{Inference.} 
Once the proposed model is trained, we employ a simple bidirectional beam search algorithm to predict the output sequence.
As illustrated in Figure~\ref{decoding},
with two special start tokens which are optimized during the training process, we let half of the beam to keep decoding from left to right, and allow the other half beam to decode from right to left.
The blue blocks denote the ongoing expansion of the hypothesis and decoding terminates when the end-of-sentence flag is predicted.
More importantly, by using the bidirectional multi-head intra-attention, the two decoding manners can help and interact with each other in one beam-search process.
Alternatively, we can also use greedy search to our model.

\begin{table*}[htbp]
	\small
	\newcommand{\tabincell}[2]{\begin{tabular}{@{}#1@{}}#2\end{tabular}}
	\centering
	\setlength{\belowcaptionskip}{-0.2cm}
	\begin{tabular}{p{3cm}|p{3cm}||cc|cc|cc}
		\hline
		\multirow{2}{*}{System} &  \multirow{2}{*}{Architecture}&   \multicolumn{2}{c}{English-German}       &   \multicolumn{2}{c}{Chinese-English}  &   \multicolumn{2}{c}{English-Romanian} \\
		\cmidrule(lr){3-4}  \cmidrule(lr){5-6}  \cmidrule(lr){7-8}
		&                 &   Quality  & Speed       &     Quality  & Speed   &   Quality       & Speed  \\
		\hline
		\hline
		\multicolumn{8}{c}{Existing NMT systems} \\
		\hline  
		\multirow{2}{*}{\cite{gu2017non}}  & NAT     &  17.35   &    N/A    &  -  &  - &   26.22   &   15.6$\times$  \\  
		& NAT (s=100)     &  19.17   &    N/A    &  -  &  - &   29.79   &   2.36$\times$  \\  
		\hline
		\multirow{2}{*}{\cite{lee2018deterministic}} & D-NAT    &   12.65    &    11.71$\times$    &  -   &   -    &   24.45     &   16.03$\times$  \\  
		& D-NAT (adaptive)    &   18.91    &    1.98$\times$    &  -   &   -    & 29.66       &   5.23$\times$  \\  
		\hline
		\multirow{2}{*}{\cite{kaiser2018fast}}& LT                &   19.80     &      3.89$\times$             &   -  &   -    & -          &  -             \\
		& LT (s=100)    &   22.50     &     N/A             &   -  &   -    & -          &  -             \\
		\hline
		\multirow{2}{*}{\tabincell{l}{\cite{wang2018semi} \\ (beam search)}}&   SAT (K=2)          &   26.90     &   1.51$\times$     &  39.57   &   1.69$\times$    & -          &  -             \\
		&   SAT (K=6)   &   24.83     &   2.98$\times$     &  35.32  &   3.18$\times$    & -          &  -             \\
		\hdashline
		\multirow{2}{*}{\tabincell{l}{\cite{wang2018semi} \\ (greedy search)}}&   SAT (K=2)          &   26.09     &   1.70$\times$     &    38.37  &   1.71$\times$    & -          &  -             \\
		&   SAT (K=6)  &   23.93     &   4.57$\times$     &   33.75  &   4.70$\times$    & -          &  -             \\
		\hline
		\hline
		\multicolumn{8}{c}{Our NMT systems} \\
		\hline
		\multirow{3}{*}{\tabincell{l}{This work \\ (beam search)}}  &  Transformer     &     27.06        &  1.00$\times$   &   46.56     &  1.00$\times$ & 32.28    & 1.00$\times$   \\  
		&  Transformer (R2L)     &    26.71    &   1.02$\times$   &   44.63    &   0.94$\times$     &   32.29     &  0.98$\times$ \\  
		&Our Model     &        \bf{27.45}        &  1.38$\times$    &  \bf{47.82} & 1.41$\times$    &  \bf{33.02}   & 1.43$\times$ \\  
		\hline
		\multirow{3}{*}{\tabincell{l}{This work \\ (greedy search)}} & Transformer     &     26.23        &  1.00$\times$   &   44.63     &  1.00$\times$ & 31.71    & 1.00$\times$   \\  
		& Transformer (R2L)   &   25.38   & 0.97$\times$   &   43.68     &  0.98$\times$ & 31.19    & 1.04$\times$   \\  
		&Our Model     &         \bf{27.22}        &  1.61$\times$    &  \bf{47.50} & 1.51$\times$    &  \bf{32.82}   & 1.46$\times$ \\  
		\hline
	\end{tabular}
	\caption{Translation quality (BLEU) and speed on official test sets. Translation speed is measured on the amount of translated sentences in one second. 
		For comparison, we also list results reported by 
		Gu {\itshape{et al}}. [2017]; Lee {\itshape{et al}}. [2018]; Kaiser {\itshape{et al}}. [2018]; Wang {\itshape{et al}}. [2018].
		Note that we and SAT use different size corpus and different preprocessing methods for Chinese-English translation. Although the non-autoregressive or semi-autoregressive NMT models have greater potential in speedup decoding than ours, the major drawback is translation quality degradation.
		By making full use of the history information and future information, our SBSG model can get a significant BLEU improvement (p $<$ 0.01) than autoregressive, semi-autoregressive and non-autoregressive models. 
	} 
	\label{mainresults}
\end{table*}

\section{Application to Neural Machine Translation}

We use BLEU~\cite{Papineni:2002} to evaluate the proposed model on translation tasks.

\subsection{Setup}

We verify our model on three translation datasets of different sizes: WMT14 English-German{\footnote[2]{http://www.statmt.org/wmt14/translation-task.html}} (En$\Rightarrow$De), NIST  Chinese-English{\footnote[3]{The corpora include LDC2000T50, LDC2002T01, LDC2002E18, LDC2003E07, LDC2003E14, LDC2003T17 and LDC2004T07.}} (Ch$\Rightarrow$En), WMT16 English-Romanian{\footnote[4]{http://www.statmt.org/wmt16/translation-task.html}} (En$\Rightarrow$Ro), whose training sets consist of 4.5M, 2.0M, 0.6M sentence pairs, respectively.
We tokenize the corpora using a script from Moses~\cite{P07-2045} and segment each word into subword units using BPE~\cite{Sennrich:2016A}.
We use 37K and 40K shared BPE tokens for En$\Rightarrow$De and En$\Rightarrow$Ro respectively.
For En$\Rightarrow$De, we use newstest2013 as the validation set and newstest2014 as the test set. 
For Ch$\Rightarrow$En, we utilize BPE to encode Chinese and English respectively, and limit the source and target vocabularies to the most frequent 30K tokens. We use NIST 2006 as the validation set, NIST 2003-2005 as our test sets.
For En$\Rightarrow$Ro, we use newsdev-2016 and newstest-2016 as development and test sets.

We implement the proposed model based on the tensor2tensor{\footnote[5]{https://github.com/tensorflow/tensor2tensor}} toolkit.
For our bidirectional Transformer model, we employ the Adam optimizer with $\beta_1$=$0.9$, $\beta_2$=$0.998$, and $\epsilon$=$10^{-9}$. We use the same warmup and decay strategy for learning rate as Vaswani et al.~\shortcite{vaswani2017attention}
, with 16,000 warmup steps. During training, we employ label smoothing of value $\epsilon_{ls}$=$0.1$. We use three GPUs to train En$\Rightarrow$De and one GPU for the other two language pairs.
 For evaluation, we use beam search with a beam size of $k$=$4$ and length penalty $\alpha$=$0.6$. Besides, we use 6 encoder and decoder layers, 512 hidden size, 8 attention-heads, 2048 feed-forward inner-layer dimensions.

\subsection{Results and Analysis}

\paragraph{Parameters.} 
NAT~\cite{gu2017non} adopts encoder-decoder architecture with additional fertility predictor model. D-NAT~\cite{lee2018deterministic} has two decoders and needs more parameters than conventional Transformer.
Our bidirectional NMT model uses one single encoder-decoder model, which can predict the target tokens in left-to-right and right-to-left manners simultaneously. 
Hence, our SBSG model does not increase any parameters except for a hyper-parameter $\lambda$ compared to the standard Transformer.

\paragraph{Inference Speed.} 
As shown in Table~\ref{mainresults}, the proposed SBSG model is capable of decoding approximately 1.4$\times$ faster than autoregressive Transformer with beam search in three translation tasks.
Besides, our model obtains 1.61$\times$ (En$\Rightarrow$De), 1.51$\times$ (Ch$\Rightarrow$En), and 1.46$\times$ (En$\Rightarrow$Ro) speedup than Transformer in greedy search.
As a compromise solution between autoregressive and non-autoregressive models, the speed of our model is relatively slower than NAT, D-NAT, and LT~\cite{kaiser2018fast}. Besides, our proposed model is capable of obtaining comparable translation speed compared to SAT~\cite{wang2018semi} with K=2.

\begin{table*}[htbp]
	\centering
	\setlength{\belowcaptionskip}{-0.3cm}
	\begin{tabular}{p{1.5cm}|p{14.5cm}}
		\hline
		Source &  \small{ 徽标\ 由\ 乘风\ 破浪\ 西行\ 的\ 帆船\ 、\ 翻滚\ 的\ 浅蓝色\ 弧形\ 水纹\ 浪花\ 和\ 文字\ "\ 郑和下西洋\ 600\ 周年\ 1405\ -\ 2005\ "\ 组成\ 。 }  \\
		\hline
		Reference   & \small{ the logo was made up of a westbound sailboat braving the wind and waves , churning arc - shaped spindrifts in light blue color , and words that say " zheng he 's 600th anniversary to the west , 1405 - 2005 . ".} \\
		\hline
		L2R &  \small{the logo is composed of sailboats that break the wind and break the waves , rolling light blue water wave flowers .  }  \\
		\hline
		R2L &  \small{ 1405-2005 of zheng he , the 600th anniversary of the west .}    \\
		\hline
		SBSG &   \small{the logo is composed of sailing boat by wind and waves , rolling light blue surpo- shaped flowers , and words " zheng to the west , 1405 - 2005 " .}  \\
		\hline
	\end{tabular}
	\caption{A Chinese-English translation example of baselines and our proposed model. 
		Our model can alleviate the unbalanced output problems [Liu {\itshape{et al}}., 2016b] by generating a sentence from both sides to the middle.} 
	\label{example}
\end{table*}

\paragraph{Translation Quality.} 
Table~\ref{mainresults} shows translation performance of En$\Rightarrow$De, Ch$\Rightarrow$En, and En$\Rightarrow$Ro translation tasks.
The proposed model behaves better than NAT, D-NAT, LT in all test datasets.  In particular, our model with beam search significantly outperforms NAT, D-NAT, and LT by 8.28, 8.54 and 4.95 BLEU points in large-scale English-German translation, respectively.
Although the SAT has a faster decoding speed than the SBSG model when K becomes bigger, it suffers from the translation quality degradation relative to the autoregressive NMT. 
Compared to autoregressive Transformer, our proposed model with beam search is able to behave better in terms of both decoding speed and translation quality.
Furthermore, our model with greedy search does not only outperform autoregressive Transformer by 0.99, 2.87 and 1.11 BLEU points of translation quality in En$\Rightarrow$De, Ch$\Rightarrow$En and En$\Rightarrow$Ro respectively, but also significantly speedups the decoding of conventional Transformer.

\begin{figure}
	\setlength{\abovecaptionskip}{+0.15cm}
	\setlength{\belowcaptionskip}{-0.2cm}
	\centering
	\includegraphics[width=8cm]{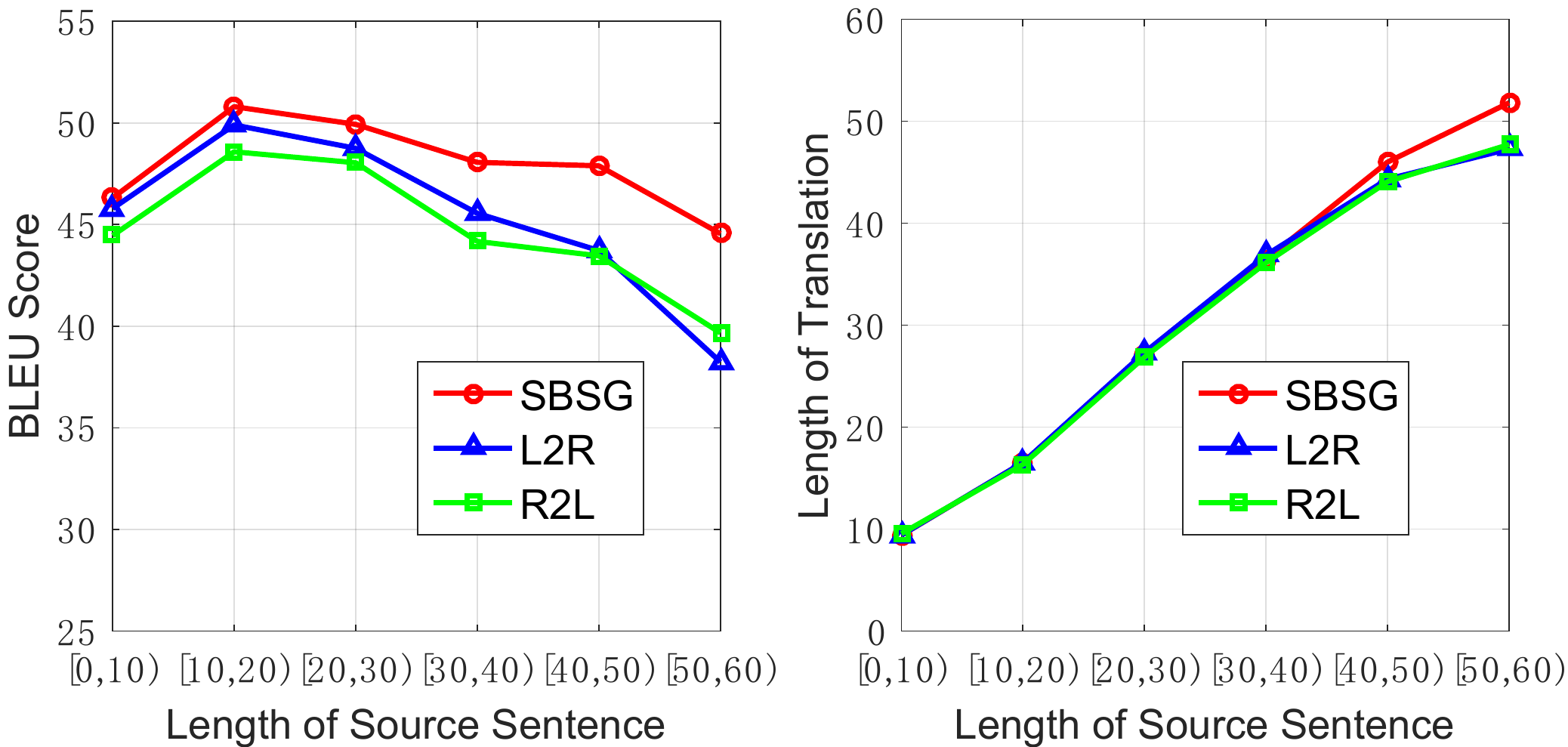}
	\caption{Length Analysis - Performance of the generated translations with respect to the lengths of the source sentences. The proposed SBSG model can alleviate under-translation by producing longer translation on long sentences.
	} \label{length}
\end{figure}

\paragraph{Length Analysis.} 
We follow ~\citeauthor{Bahdanau:2015}~\shortcite{Bahdanau:2015} to group sentences of similar lengths together, and compute a BLEU score and the averaged length of translations per group. 
Figure ~\ref{length} shows that the performance of Transformer and Transformer (R2L) drops rapidly when the length of the input sentence increases. 
Our SBSG model alleviates this problem by generating a sequence from both sides to the middle, which in general encourages the model to produce more accurate and long sentences.

\paragraph{Case Study.} 
In Table~\ref{example}, we present a translation example from NIST Chinese-English. \citeauthor{liu2016agreementa}~\shortcite{liu2016agreementa} show that L2R model produces the translation with good prefix and sometimes omits the second half sentence, but L2R model usually generates the generation with better suffixes. Our results confirm these findings. The proposed SBSG model can alleviate the errors by generating sequences from both sides to the middle.

\section{Application to Text Summarization}

We further verify the effectiveness of our proposed SBSG model on text summarization, which is another real-world application that encoder-decoder framework succeeds~\cite{D15-1044}.

\subsection{Setup}

Abstractive sentence summarization aims to provide a title-like summary for a long sentence. We conduct text summarization experiments on English Gigaword dataset{\footnote[6]{https://github.com/harvardnlp/sent-summary}}.
The parallel corpus is produced by pairing the first sentence and the headline in the news article with some heuristic rules.
The extracted corpus contains about 3.8M sentence-summary pairs for the training set and 189K examples for the development set. 
We employ a shared vocabulary of about 90K word types and use DUC 2004 used by \citeauthor{D15-1044}~\shortcite{D15-1044} as our test set. 
The model structure is the same as that used in neural machine translation.
We employ ROUGE as our evaluation metric, which is all widely adopted evaluation metric for text summarization.

\begin{table}[htbp]
	\small
	\centering
	\setlength{\belowcaptionskip}{-0.2cm}
	\begin{tabular}{lcccc}
		\hline
		\textbf{DUC2004}              &    RG-1 &  RG-2     &    RG-L    & Speed   \\
		\hline
		ABS$\ddagger$              &      26.55       &   7.06    &   22.05  & -      \\
		Feats2s$\ddagger$               &      28.35       &   9.46    &   24.59  &  -   \\
		Selective-Enc$\ddagger$               &      29.21       &   9.56    &   25.51  &  -   \\
		Transformer                &      28.09       &   9.52    &  24.91  &  1.00$\times$ \\
		SBSG (beam)          &      28.77      &   10.11    &       26.11  &   1.48$\times$ \\
		SBSG (greedy)          &      28.70      &   9.88    &       25.93  &   2.09$\times$ \\
		\hline
	\end{tabular}
	\caption{ROUGE recall evaluation results on DUC 2004 test set.
		For comparison, we also list results reported by Rush {\itshape{et al}}. [2015]; Nallapati {\itshape{et al}}. [2016]; Zhou {\itshape{et al}}. [2017].
		Results with $\ddagger$ mark are taken from the corresponding papers.
		Our proposed SBSG model significant outperforms the conventional Transformer model in terms of both decoding speed and generation quality.
	} 
	\label{sumarization}
\end{table}

\subsection{Results and Analysis}

In Table~\ref{sumarization}, we report the ROUGE score and speed 
for DUC 2004 test set.
Experiments show that the generation quality of our proposed model is on par with the state-of-the-art text summarization models.
We observe approximately 1.5$\times$ faster decoding than the autoregressive Transformer while achieving better generation quality. Specially, our model with beam search (greedy search) is capable of decoding 1.64$\times$ (2.26$\times$) faster than conventional Transformer on English Gigaword test set.

\section{Conclusions}

In this work, we propose a novel SBSG model that performs bidirectional decoding simultaneously and interactively.
Instead of producing output sentence token by token, the proposed model makes decoding much more parallelizable and generates two tokens at each time step.
We extensively evaluate the proposed SBSG model on neural machine translation (En$\Rightarrow$De, Ch$\Rightarrow$En, and En$\Rightarrow$Ro) and text summarization (English Gigaword) tasks. Different from previous non-autoregressive models~\cite{gu2017non,lee2018deterministic,kaiser2018fast} which suffer from serious quality degradation,  our SBSG model achieves a significant improvement in both generation quality and decoding speed compared to the state-of-the-art autoregressive Transformer.

\section*{Acknowledgments}

The research work described in this paper has been funded by the National Key Research and Development Program of China under Grant No. 2016QY02D0303 and the Natural Science Foundation of China under Grant No. U1836221 and 61673380.
This work is also supported by grants from NVIDIA NVAIL program.

\bibliographystyle{named}
\bibliography{ijcai19}

\begin{thebibliography}{}

\bibitem[\protect\citeauthoryear{Bahdanau \bgroup \em et al.\egroup
  }{2015}]{Bahdanau:2015}
Dzmitry Bahdanau, Kyunghyun Cho, and Yoshua Bengio.
\newblock Neural machine translation by jointly learning to align and
  translate.
\newblock {\em In ICLR}, 2015.

\bibitem[\protect\citeauthoryear{Devlin}{2017}]{D17-1300}
Jacob Devlin.
\newblock Sharp models on dull hardware: Fast and accurate neural machine
  translation decoding on the cpu.
\newblock In {\em EMNLP}, pages 2820--2825, 2017.

\bibitem[\protect\citeauthoryear{Finch and Sumita}{2009}]{D09-1117}
Andrew Finch and Eiichiro Sumita.
\newblock Bidirectional phrase-based statistical machine translation.
\newblock In {\em EMNLP}, pages 1124--1132, 2009.

\bibitem[\protect\citeauthoryear{Gehring \bgroup \em et al.\egroup
  }{2017}]{gehring2017convolutional}
Jonas Gehring, Michael Auli, David Grangier, Denis Yarats, and Yann Dauphin.
\newblock Convolutional sequence to sequence learning.
\newblock In {\em ICML}, 2017.

\bibitem[\protect\citeauthoryear{Gu \bgroup \em et al.\egroup
  }{2017}]{gu2017non}
Jiatao Gu, James Bradbury, Caiming Xiong, Victor~OK Li, and Richard Socher.
\newblock Non-autoregressive neural machine translation.
\newblock {\em arXiv preprint arXiv:1711.02281}, 2017.

\bibitem[\protect\citeauthoryear{Kaiser \bgroup \em et al.\egroup
  }{2018}]{kaiser2018fast}
{\L}ukasz Kaiser, Aurko Roy, Ashish Vaswani, Niki Pamar, Samy Bengio, Jakob
  Uszkoreit, and Noam Shazeer.
\newblock Fast decoding in sequence models using discrete latent variables.
\newblock {\em arXiv preprint arXiv:1803.03382}, 2018.

\bibitem[\protect\citeauthoryear{Kim and Rush}{2016}]{kim2016sequence}
Yoon Kim and Alexander~M. Rush.
\newblock Sequence-level knowledge distillation.
\newblock In {\em EMNLP}, 2016.

\bibitem[\protect\citeauthoryear{Koehn \bgroup \em et al.\egroup
  }{2007}]{P07-2045}
Philipp Koehn, Hieu Hoang, Alexandra Birch, Chris Callison-Burch, Marcello
  Federico, Nicola Bertoldi, Brooke Cowan, Wade Shen, Christine Moran, Richard
  Zens, Chris Dyer, Ondrej Bojar, Alexandra Constantin, and Evan Herbst.
\newblock Moses: Open source toolkit for statistical machine translation.
\newblock In {\em ACL}, 2007.

\bibitem[\protect\citeauthoryear{Lee \bgroup \em et al.\egroup
  }{2018}]{lee2018deterministic}
Jason Lee, Elman Mansimov, and Kyunghyun Cho.
\newblock Deterministic non-autoregressive neural sequence modeling by
  iterative refinement.
\newblock In {\em EMNLP}, pages 1173--1182, 2018.

\bibitem[\protect\citeauthoryear{Li \bgroup \em et al.\egroup
  }{2018}]{li-etal-2018-ensure}
Haoran Li, Junnan Zhu, Jiajun Zhang, and Chengqing Zong.
\newblock Ensure the correctness of the summary: Incorporate entailment
  knowledge into abstractive sentence summarization.
\newblock In {\em COLING}, 2018.

\bibitem[\protect\citeauthoryear{Liu \bgroup \em et al.\egroup
  }{2016a}]{liu2016agreementb}
Lemao Liu, Andrew~M Finch, Masao Utiyama, and Eiichiro Sumita.
\newblock Agreement on target-bidirectional lstms for sequence-to-sequence
  learning.
\newblock In {\em AAAI}, 2016.

\bibitem[\protect\citeauthoryear{Liu \bgroup \em et al.\egroup
  }{2016b}]{liu2016agreementa}
Lemao Liu, Masao Utiyama, Andrew Finch, and Eiichiro Sumita.
\newblock Agreement on target-bidirectional neural machine translation.
\newblock In {\em NAACL}, 2016.

\bibitem[\protect\citeauthoryear{Liu \bgroup \em et al.\egroup
  }{2018}]{liu2018}
Yuchen Liu, Long Zhou, Yining Wang, Yang Zhao, Jiajun Zhang, and Chengqing
  Zong.
\newblock A comparable study on model averaging, ensembling and reranking in
  nmt.
\newblock In {\em NLPCC}, pages 299--308, 2018.

\bibitem[\protect\citeauthoryear{Mi \bgroup \em et al.\egroup
  }{2016}]{P16-2021}
Haitao Mi, Zhiguo Wang, and Abe Ittycheriah.
\newblock Vocabulary manipulation for neural machine translation.
\newblock In {\em ACL}, pages 124--129, 2016.

\bibitem[\protect\citeauthoryear{Oord \bgroup \em et al.\egroup
  }{2017}]{oord2017parallel}
Aaron van~den Oord, Yazhe Li, Igor Babuschkin, Karen Simonyan, Oriol Vinyals,
  Koray Kavukcuoglu, George van~den Driessche, Edward Lockhart, Luis~C Cobo,
  Florian Stimberg, et~al.
\newblock Parallel wavenet: Fast high-fidelity speech synthesis.
\newblock {\em arXiv preprint arXiv:1711.10433}, 2017.

\bibitem[\protect\citeauthoryear{Papineni \bgroup \em et al.\egroup
  }{2002}]{Papineni:2002}
Kishore Papineni, Salim Roukos, Todd Ward, and WeiJing Zhu.
\newblock Bleu: a methof for automatic evaluation of machine translation.
\newblock In {\em ACL}, 2002.

\bibitem[\protect\citeauthoryear{Rush \bgroup \em et al.\egroup
  }{2015}]{D15-1044}
Alexander~M. Rush, Sumit Chopra, and Jason Weston.
\newblock A neural attention model for abstractive sentence summarization.
\newblock In {\em EMNLP}, 2015.

\bibitem[\protect\citeauthoryear{Sennrich \bgroup \em et al.\egroup
  }{2016a}]{W16-2323}
Rico Sennrich, Barry Haddow, and Alexandra Birch.
\newblock Edinburgh neural machine translation systems for wmt 16.
\newblock In {\em WMT}, 2016.

\bibitem[\protect\citeauthoryear{Sennrich \bgroup \em et al.\egroup
  }{2016b}]{Sennrich:2016A}
Rico Sennrich, Barry Haddow, and Alexandra Birch.
\newblock Neural machine translation of rare words with subword units.
\newblock In {\em ACL}, pages 1715--1725, 2016.

\bibitem[\protect\citeauthoryear{Serdyuk \bgroup \em et al.\egroup
  }{2018}]{serdyuk2018twin}
Dmitriy Serdyuk, Nan~Rosemary Ke, Alessandro Sordoni, Adam Trischler, Chris
  Pal, and Yoshua Bengio.
\newblock Twin networks: Matching the future for sequence generation.
\newblock In {\em ICLR}, 2018.

\bibitem[\protect\citeauthoryear{Sutskever \bgroup \em et al.\egroup
  }{2014}]{Sutskever:2014}
Ilya Sutskever, Oriol Vinyals, and Quoc~VV Le.
\newblock Sequence to sequence learning with neural networks.
\newblock In {\em NIPS}, pages 3104--3112, 2014.

\bibitem[\protect\citeauthoryear{Vaswani \bgroup \em et al.\egroup
  }{2017}]{vaswani2017attention}
Ashish Vaswani, Noam Shazeer, Niki Parmar, Jakob Uszkoreit, Llion Jones,
  Aidan~N Gomez, {\L}ukasz Kaiser, and Illia Polosukhin.
\newblock Attention is all you need.
\newblock In {\em NIPS}, pages 5998--6008, 2017.

\bibitem[\protect\citeauthoryear{Vinyals \bgroup \em et al.\egroup
  }{2015}]{Vinyals2015Show}
Oriol Vinyals, Alexander Toshev, Samy Bengio, and Dumitru Erhan.
\newblock Show and tell: A neural image caption generator.
\newblock In {\em CVPR}, 2015.

\bibitem[\protect\citeauthoryear{Wang \bgroup \em et al.\egroup
  }{2018}]{wang2018semi}
Chunqi Wang, Ji~Zhang, and Haiqing Chen.
\newblock Semi-autoregressive neural machine translation.
\newblock {\em arXiv preprint arXiv:1808.08583}, 2018.

\bibitem[\protect\citeauthoryear{Watanabe and
  Sumita}{2002}]{watanabe2002bidirectional}
Taro Watanabe and Eiichiro Sumita.
\newblock Bidirectional decoding for statistical machine translation.
\newblock In {\em COLING}, 2002.

\bibitem[\protect\citeauthoryear{Xu \bgroup \em et al.\egroup
  }{2015}]{Xu2015Show}
Kelvin Xu, Jimmy Ba, Ryan Kiros, Kyunghyun Cho, Aaron Courville, Ruslan
  Salakhutdinov, Richard Zemel, and Yoshua Bengio.
\newblock Show, attend and tell: Neural image caption generation with visual
  attention.
\newblock {\em Computer Science}, pages 2048--2057, 2015.

\bibitem[\protect\citeauthoryear{Zhang \bgroup \em et al.\egroup
  }{2018a}]{P18-1166}
Biao Zhang, Deyi Xiong, and jinsong Su.
\newblock Accelerating neural transformer via an average attention network.
\newblock In {\em ACL}, pages 1789--1798, 2018.

\bibitem[\protect\citeauthoryear{Zhang \bgroup \em et al.\egroup
  }{2018b}]{zhang2018asynchronous}
Xiangwen Zhang, Jinsong Su, Yue Qin, Yang Liu, Rongrong Ji, and Hongji Wang.
\newblock Asynchronous bidirectional decoding for neural machine translation.
\newblock {\em In AAAI}, 2018.

\bibitem[\protect\citeauthoryear{Zhou \bgroup \em et al.\egroup
  }{2017}]{P17-1101}
Qingyu Zhou, Nan Yang, Furu Wei, and Ming Zhou.
\newblock Selective encoding for abstractive sentence summarization.
\newblock In {\em ACL}, pages 1095--1104, 2017.

\bibitem[\protect\citeauthoryear{Zhou \bgroup \em et al.\egroup
  }{2019}]{zhou2019synchronous}
Long Zhou, Jiajun Zhang, and Chengqing Zong.
\newblock Synchronous bidirectional neural machine translation.
\newblock In {\em TACL}, pages 91--105, 2019.

\end{thebibliography}

\end{CJK*}
\end{document}